\documentclass[11pt,letterpaper]{article}
\usepackage{naaclhlt2016}
\usepackage{times}
\usepackage{latexsym}
\usepackage{graphicx}

\naaclfinalcopy % Uncomment this line for the final submission
\def\naaclpaperid{177} %  Enter the naacl Paper ID here

\title{Bidirectional RNN for Medical Event Detection in Electronic Health Records}

\author{Abhyuday N Jagannatha$^1$, Hong Yu$^{1,2}$\\
$^1$ University of Massachusetts, MA, USA\\
$^2$ Bedford VAMC and CHOIR, MA, USA\\
{\tt abhyuday@cs.umass.edu} , {\tt hong.yu@umassmed.edu}\\
}

\date{}

\begin{document}

\maketitle

%\begin{abstract}
%Sequence labelling for extraction of medical events and their attributes from unstructured text in Electronic Health Record (EHR) notes is a key step towards semantic understanding of EHRs. It has varied applications in health informatics including pharma-co-vigilance and drug surveillance. Most of the current supervised machine learning models in this domain use Conditional Random Fields (CRFs). In this application, we explored recurrent neural network frameworks and show that they significantly outperformed Conditional Random Field based labeling models.
%\end{abstract}

\begin{abstract}
Sequence labeling for extraction of medical events and their attributes from unstructured text in Electronic Health Record (EHR) notes is a key step towards semantic understanding of EHRs. It has important applications in health informatics including pharmacovigilance and drug surveillance. The state of the art supervised machine learning models in this domain are based on Conditional Random Fields (CRFs) with features calculated from fixed context windows. In this application, we explored recurrent neural network frameworks\footnote{RNN Code is available at https://github.com/abhyudaynj/birnn-bionlp} and show that they significantly outperformed the CRF models.
\end{abstract}

\section{Introduction}
EHRs report patient's health, medical history and treatments compiled by medical staff at hospitals. It is well known that EHR notes contain information about medical events including medication, diagnosis (or Indication), and adverse drug events (ADEs) etc. A medical event in this context can be described as a change in patient's medical status. Identifying these events in a structured manner has many important clinical applications such as discovery of abnormally high rate of adverse reaction events to a particular drug, surveillance of drug efficacy, etc. In this paper we treat EHR clinical event detection as a task of sequence labeling.

Sequence labeling in the context of machine learning refers to the task of learning to predict a label for each data-point in a sequence of data-points. This learning framework has wide applications in many disciplines such as genomics, intrusion detection, natural language processing, speech recognition etc. However, sequence labeling in EHRs is a challenging task. Unlike text in the open domain, EHR notes are frequently noisy, containing incomplete sentences, phrases and irregular use of language. In addition, EHR notes incorporate abundant abbreviations, rich medical jargons, and their variations, which make recognizing semantically similar patterns in EHR notes difficult. Additionally, different events exhibit different patterns and possess different prevalences. For example, while a medication comprises of at most a few words of a noun, an ADE (e.g., ``has not felt back to his normal self'') may vary to comprise of a significant part of a sentence. While medication information is frequently described in EHRs, ADEs are typically rare events. 

Rule-based and learning-based approaches have been developed to identify and extract information from EHR notes ~\cite{haerian_methods_2012},~\cite{xu_medex:_2010},~\cite{friedman_general_1994},~\cite{aronson_effective_2001},~\cite{polepalli_ramesh_automatically_2014}. Learning-based approaches use sequence labeling algorithms like Conditional Random Fields ~\cite{lafferty_conditional_2001}, Hidden Markov Models ~\cite{collier_extracting_2000}, and Max-entropy Markov Models ~\cite{mccallum_maximum_2000}. One major drawback of these graphical models is that the label prediction at any time point only depends on its data instance and the immediate neighboring labels.

While this approach performs well in learning the distribution of the output labels, it has some limitations. One major limitation is that it is not designed to learn from dependencies which lie in the surrounding but not quite immediate neighborhood. Therefore, the feature vectors have to be explicitly modeled to include the surrounding contextual information. Traditionally, bag of words representation of surrounding context has shown reasonably good performance. However, the information contained in the bag of words vector is very sensitive to context window size. If the context window is too short, it will not include all the information. On the other hand if the context window is too large, it will compress the vital information with other irrelevant words. Usually a way to tackle this problem is to try different context window sizes and use the one that gives the highest validation performance. However, this method cannot be easily applied to our task, because different medical events like medication, diagnosis or adverse drug reaction require different context window sizes. For example, while a medication can be determined by a context of two or three words containing the drug name, an adverse drug reaction would require the context of the entire sentence.  As an example, this is a sentence from one of the EHRs, ``The follow-up needle biopsy results were consistent with \textit{bronchiolitis obliterans}, which was likely due to the Bleomycin component of his \textit{ABVD chemo}''. In this sentence, the true labels are \textit{Adverse Drug Event(ADE)} for ``\textit{bronchiolitis obliterans}''  and \textit{Drugname} for ``\textit{ABVD chemo}''. However the ADE , ``\textit{bronchiolitis obliterans}'' could be miss-labeled as just another disease or symptom, if the entire sentence is not taken into context.

Recent advancements in Recurrent Neural Networks (RNNs) have opened up new avenues of research in sequence labeling. Traditionally, recurrent neural networks have been hard to train through Back-Propagation, because learning long term dependencies using simple recurrent neurons lead to problems like exploding or vanishing gradients ~\cite{bengio_learning_1994}, ~\cite{hochreiter_gradient_2001}. Recent approaches have modified the simple neuron structure in order to learn dependencies over longer intervals more efficiently. In this study, we evaluate the performance of two such neural networks, namely, Long Short Term Memory (LSTM) and Gated Recurrent Units (GRU).

Timely identification of new drug toxicities is an unresolved clinical and public health problem, costing people's lives and billions of US dollars. In this study, we empirically evaluated LSTM and GRU on EHR notes, focusing on the clinically important task of detecting medication, diagnosis, and adverse drug event. To our knowledge, we are the first group reporting the uses of RNN frameworks for information extraction in EHR notes. 

\section{Related Work}
Medication and ADE detection is an important NLP task in biomedicine. Related existing NLP approaches can be grouped into knowledge or rule-based, supervised machine learning, and hybrid approaches. For example, Hazlehurst et al. ~\shortcite{hazlehurst_mediclass:_2005} developed MediClass, a knowledge-based system that deploys a set of domain-specific logical rules for medical concept extraction. Wang et al. ~\shortcite{wang_medical_2015} , Humphreys et.al. ~\shortcite{da_unified_1993-1} and others map EHR notes to medical concepts to an external knowledge resource using hybrid rule-based and syntactic parsing approaches. Gurulingappa et al. ~\shortcite{gurulingappa_empirical_2010} detect two medical entities (\textit{disease} and \textit{adverse events}) in a corpus of annotated Medline abstracts. In contrast, our work uses a corpus of actual medical notes and detects additional events and attributes.

Rochefort et al. ~\shortcite{rochefort_novel_2015} developed document classifiers to classify whether a clinical note contains deep venous thromboembolisms and pulmonary embolism. Haerian et al. ~\shortcite{haerian_methods_2012} applied distance supervision to identify terms (e.g., including ``suicidal'', ``self harm'', and ``diphenhydramine overdose'') associated with suicide events. Zuofeng Li et al. ~\shortcite{li_lancet:_2010} extracted medication information using CRFs.

Many named entity recognition systems in the biomedical domain have been driven by the Shared tasks of BioNLP ~\cite{kim_overview_2009},  BioCreAtivE ~\cite{hirschman_overview_2005} i2b2 shared NLP tasks ~\cite{li_extracting_2009} and ShARe/CLEF evaluation tasks ~\cite{pradhan_evaluating_2014}. The best performing clinical NLP systems for named entity recognition includes Tang et al ~\shortcite{tang_recognizing_2013} which applied CRF and structured SVM.

Neural Network models like Convolutional Neural Networks and Recurrent Neural Networks (LSTM, GRU) have recently been been successfully used to tackle various sequence labeling problems in NLP. Collobert ~\shortcite{collobert_natural_2011} used Convolutional Neural Network for sequence labeling problems like POS tagging, NER etc. . Later, Huang et al. ~\shortcite{huang_bidirectional_2015} achieved comparable or better scores using bi-directional LSTM based models.

\section{Dataset}
\begin{table}
\small
\centering
\begin{tabular}{| l | l | p{2cm} |}
\hline \bf Labels & \bf Annotations & \bf Avg. Words / Annotations \\ \hline \hline
ADE&905&1.51 \\ \hline
Indication&1988&2.34 \\ \hline
Other SSD&26013&2.14 \\ \hline
Severity &1928 &1.38 \\ \hline
Drugname&9917&1.20 \\ \hline
Duration&562&2.17 \\ \hline
Dosage&3284&2.14 \\ \hline
Route&1810&1.14 \\ \hline
Frequency&2801&2.35 \\ \hline
\end{tabular}
\caption{\label{data-table} Annotation statistics for the corpus.}
\end{table}

The annotated corpus contains 780 English EHR notes or 613,593 word tokens (an average of 786 words per note) from cancer patients who have been diagnosed with hematological malignancy. Each note was annotated by at least two annotators with inter-annotator agreement of 0.93 kappa. The annotated events and attributes and their instances in the annotated corpus are shown in Table 1. 

The annotated events can be broadly divided into two groups, \textit{Medication}, and \textit{Disease}. The \textit{Medication} group contains \textit{Drugname}, \textit{Dosage}, \textit{Frequency}, \textit{Duration} and \textit{Route}. It corresponds to information about medication events and their attributes. The attributes (Route, Frequency, Dosage, and Duration) of a medication (Drug name) occur less frequently than the Drugname tag itself, because few EHRs report complete attributes of an event.  

The \textit{Disease} group contains events related to diseases (\textit{ADE}, \textit{Indication}, \textit{Other SSD}) and their attributes (\textit{Severity}). An injury or disease can be labeled as \textit{ADE}, \textit{Indication}, or \textit{Other} \textit{SSD} depending on the semantic context. It is marked as \textit{ADE} if it is the side effect of a drug. It is marked as \textit{Indication} if it is being diagnosed currently by the doctor and a medication has been prescribed for it. Any sign, symptom or disease that does not fall into the aforementioned two categories is labeled as \textit{Other SSD}.  \textit{Other SSD} is the most common label in our corpus, because it is frequently used to label conditions in the past history of the patient.

For each note, we removed special characters that do not serve as punctuation and then split the note into sentences using regular expressions.

\section{Methods}

\subsection{Long Short Term Memory}
Long Short Term Memory Networks ~\cite{hochreiter_long_1997} are a type of Recurrent Neural Networks (RNNs). RNNs are modifications of feed-forward neural networks with recurrent connections. In a typical NN, the neuron  output at time $t$ is given by:
\begin{equation}
y_i^t=\sigma(W_ix_t+b_i)
\end{equation}
Where $W_i$ is the weight matrix, $b_i$ is the bias term and $\sigma$ is the sigmoid activation function. In an RNN, the output of the neuron at time $t-1$ is fed back into the neuron. The new activation function now becomes:
\begin{equation}
y_i^t=\sigma(W_ix_t+U_i y_i^{t-1}+b_i)
\end{equation}
Since these RNNs use the previous outputs as recurrent connections, their current output depends on the previous states. This property remembers previous information about the sequence, making them useful for sequence labeling tasks.  RNNs can be trained through back-propagation through time. Bengio et al. ~\shortcite{bengio_learning_1994} showed that learning long term dependencies in recurrent neural networks through gradient decent is difficult. This is mainly because the back-propagating error can frequently ``blow-up'' or explode which makes convergence infeasible, or it can vanish which renders the network incapable of learning long term dependencies~\cite{hochreiter_gradient_2001}.

In contrast, LSTM networks were proposed as solutions for the vanishing gradient problem and were designed to efficiently learn long term dependencies. LSTMs accomplish this by keeping an internal state that represents the memory cell of the LSTM neuron. This internal state can only be read and written through gates which control the information flowing through the cell state. The updates of various gates can be computed as:
\begin{equation}
i_t=tanh(W_{xi} x_t+W_{hi} h_{t-1})
\end{equation}
\begin{equation}
f_t=\sigma(W_{xf} x_t+W_{hf} h_{t-1})
\end{equation}
\begin{equation}
o_t=\sigma(W_{xo} x_t+W_{ho} h_{t-1})
\end{equation}
Here $i_t$ , $f_t$ and $o_t$ denote input, forget and output gate respectively. The forget and input gate determine the contributions of the previous output and the current input, in the new cell state $c_t$. The output gate controls how much of  $c_t$ is exposed as the output. The new cell state $c_t$ and the output $h_t$ can be calculated as follows:
\begin{equation}
c_t=f_t \odot c_{t-1} +i_t \odot tanh(W_{xc} x_t+W_{hc} h_{t-1})
\end{equation}
\begin{equation}
h_t=o_t \odot tanh(c_t)
\end{equation}

\begin{figure}[t]
\begin{center}
   \includegraphics[width=0.47\textwidth]{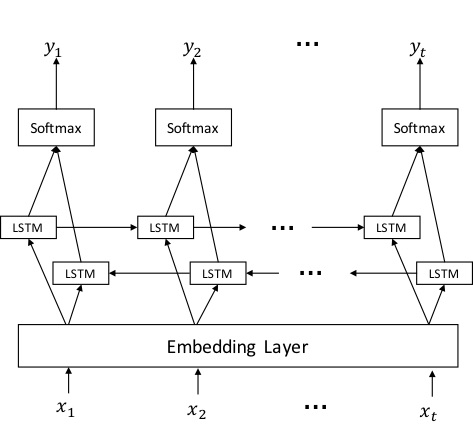}
\end{center}
   \caption{Sequence Labeling model for LSTM network}
\label{fig:lstm-model}
\end{figure}

The cell state stores relevant information from the previous time-steps. It can only be modified in an additive fashion via the input and forget gates. Simplistically, this can be viewed as allowing the error to flow back through the cell state unchecked till it back propagates to the time-step that added the relevant information. This nature allows LSTM to learn long term dependencies. 

We use LSTM cells in the Neural Network setup shown in figure 1. Here $x_k$,$y_k$  are the input word, and the predicted label for the $k^{th}$ word in the sentence. The embedding layer contains the word vector mapping from words to dense n-dimensional vector representations. We initialize the embedding layer at the start of the training with word vectors calculated on the larger data corpus described in section 4.4. This ensures that words which are not seen frequently in the labeled data corpus still have a reasonable vector representation. This step is necessary because our unlabeled corpus is much larger than the labeled one.

The words are mapped into their corresponding vector representations and fed into the LSTM layer. The LSTM layer consists of two LSTM chains, one propagating in the forward direction and other in the backward direction. We concatenate the output from the two chains to form a combined representation of the word and its context. This concatenated vector is then fed into a feed-forward neuron with Softmax activation function. The Softmax activation function normalizes the outputs to produce probability like outputs for each label type $j$ as follows:
\begin{equation}
P(l_t=j|u_t)=\frac{exp(u_t W_j)}{\sum_{k=1}^Kexp(u_t W_k)}
\end{equation}
Here $l_t$  and $u_t$ are the label and the concatenated vector for  each time step $t$. The most likely label at each word position is selected. The entire network is trained through back-propagation. The embedding vectors are also updated based on the back-propagated errors.

\subsection{Gated Recurrent Units}
Gated Recurrent Unit (GRU) is another type of recurrent neural network which was recently proposed for the purposes of Machine Translation by Cho et. al. ~\shortcite{cho_properties_2014}. Similar to LSTMs, Gated Recurrent Units also have an additive mechanism to update the cell state, with the current update. However, GRUs have a different mechanism to create the update. The candidate activation $\widetilde{h_t}$ is computed based on the previous cell state  and the current input .
\begin{equation}
\widetilde{h_t}=\sigma(W_{xh} x_t+W_{hh} (r_t \odot h_{t-1} ))
\end{equation}
Here $r_t$ is the reset gate and it controls the use of previous cell state while calculating the input activation. The reset gate itself is also computed based on the previous cell activation $h_{t-1}$ and the current candidate activation .
\begin{equation}
\widetilde{r_t}=\sigma(W_{xr} x_t+W_{hr} h_{t-1})
\end{equation}
The current cell state or activation is a linear combination of previous cell activation and the candidate activation.
\begin{equation}
h_t=(1-z_t) \odot h_{t-1}+z_t \odot \widetilde h_t
\end{equation}
Here, $z_t$ is the update gate which decides how much contribution the candidate activation and the previous cell state should have in the cell activation. The update gate is computed using the following equation:
\begin{equation}
z_t=σ(W_{hz} h_{t-1}+W_{xz} x_t)
\end{equation}
Gated recurrent units have some fundamental differences with LSTM. For example, there is no mechanism like the output gate which controls the exposure of the cell activation, instead the entire current cell activation is used as output. The mechanisms for using the previous output for the calculation of the current activation are also very different. Recent experiments ~\cite{chung_empirical_2014}, ~\cite{jozefowicz_empirical_2015} comparing both these architectures have shown GRUs to have comparable or sometimes better performance than LSTM in several tasks with long term dependencies.

We use GRU with the same Neural Network structure as shown in Figure 1 by replacing the LSTM nodes with GRU. The embedding layer used here is also initialized in a similar fashion as the LSTM network.

\begin{table}
\small
\centering
\begin{tabular}{| l | l | l | l |}
\hline \bf Models & \bf Recall & \bf Precision & \bf F-score \\ \hline \hline
CRF-nocontext&0.6562&0.7330&0.6925\\ \hline
CRF-context&0.6806&0.7711&0.7230\\ \hline
LSTM-sentence&0.8024&0.7803&0.7912\\ \hline
GRU-sentence&0.8013&0.7802&0.7906\\ \hline
LSTM-document&0.8050&0.7796&0.7921\\ \hline
GRU-document& \bf 0.8126 & \bf 0.7938 & \bf 0.8031\\ \hline
\end{tabular}
\caption{\label{results-table} Cross validated micro-average of Precision, Recall and F-score for all medical tags}
\end{table}

\subsection{The Baseline System}
CRFs have been widely used for sequence labeling tasks in NLP. CRFs model the complex dependence of the outputs in a sequence using Probabilistic Graphical Models. Probabilistic Graphical Models represent relationships between variables through a product of factors where each factor is only influenced by a smaller subset of the variables. A particular factorization of the variables provides a specific set of independence relations enforced on the data. Unlike Hidden Markov Models which model the joint $p(x,y)$, CRFs model the posterior probability $p(y|x)$ directly. The conditional can be written as a product of factors as follows:
\begin{equation}
p(y|x)=\frac{1}{Z(x)} \prod_{t=1}^T \psi_t (y_t,y_{t-1},x_t)
\end{equation}
Here $Z$ is the partition function used for normalization, $\psi_t$ are the local factor functions.

CRFs are fed the word inputs and their corresponding skip-gram word embedding ( section 4.4). To compare CRFs with RNN, we add extra context feature for each word. This is done because our aim is to show that RNNs perform better than CRFs using context windows. This extra feature consists of two vectors that are bag of words representation of the sentence sections before and after the word respectively. We add this feature to explicitly provide a mechanism that is somewhat similar to the surrounding context that is generated in a Bi-directional RNN as shown in Figure 1. This CRF model is referred to as CRF-context in our paper. We also evaluate a CRF-nocontext model, which trains a CRF without the context features. 

The tagging scheme used with both CRF models is BIO (Begin, Inside and Outside). We did not use the more detailed BILOU scheme (Begin, Inside, Last, Outside, Unit) due to data sparsity in some of the rarer labels.  

\begin{figure}[t]
\begin{center}
   \includegraphics[width=0.5\textwidth]{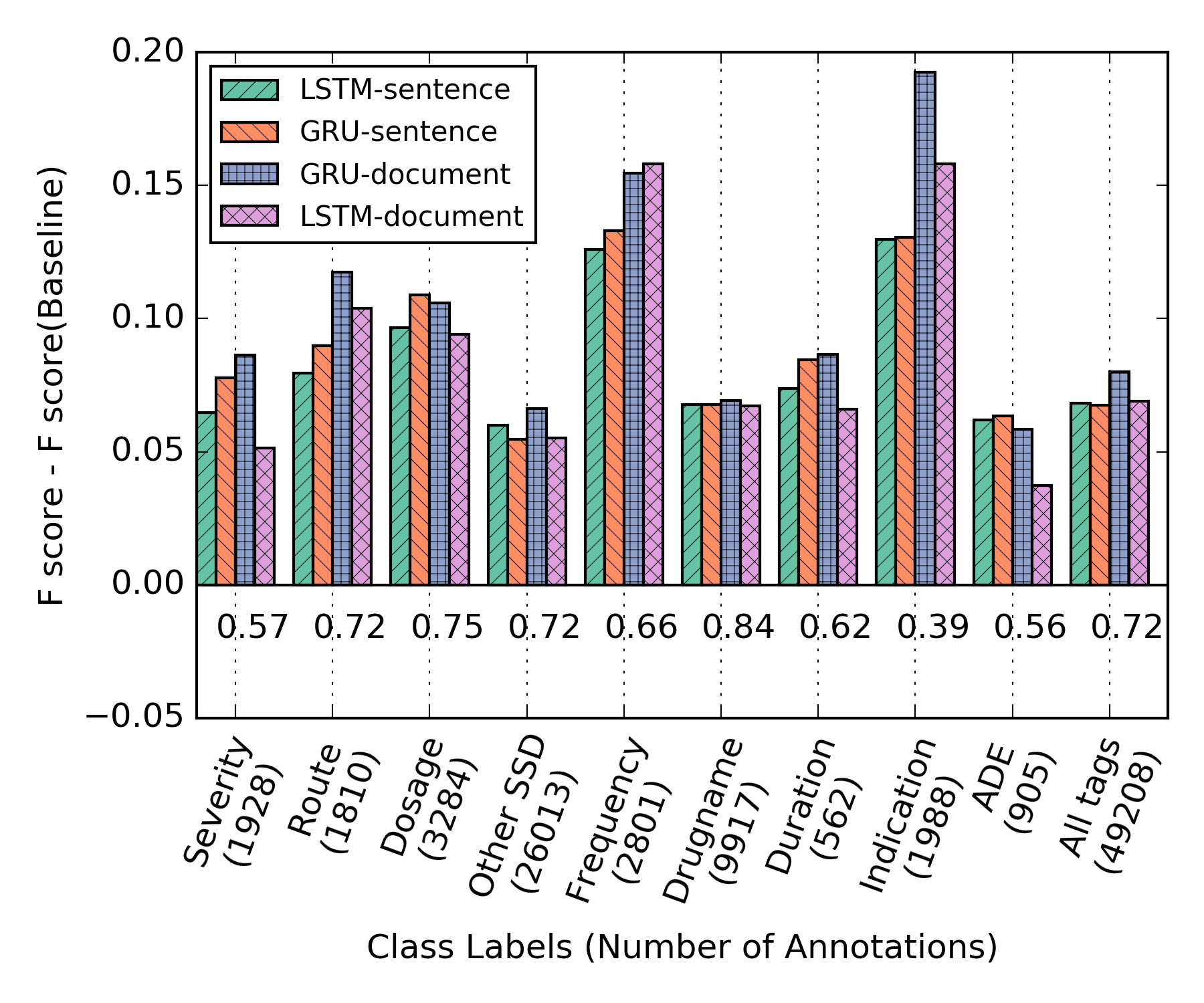}
\end{center}
   \caption{Change in F-score for RNN models with respect to CRF-context (baseline). The values below the plotted bars represent the baseline f-scores for each class label.}
\label{fig:score-change}
\end{figure}

\subsection{Skip-Gram Word Embeddings}
We use skip-gram word embeddings trained through a shallow neural network as shown by Mikolov et al.,  ~\shortcite{mikolov_distributed_2013} to initialize the embedding layer of the RNNs. This embedding is also used in the baseline CRF model as a feature. The embeddings are trained on a large unlabeled biomedical dataset, compiled from three sources, the English Wikipedia, an unlabeled EHR corpus, and PubMed Open Access articles. The English Wikipedia consists of text extracted from all the articles of English Wikipedia 2015. The unlabeled EHR corpus contains 99,700 electronic health record notes. PubMed Open Access articles are obtained by extracting the raw text from all openly available PubMed articles. This combined raw text corpus contains more than 3 billion word tokens. We convert all words to lowercase and use a context window of 10 words to train a 200 dimensional skip gram word embedding.

\section{Experiments and Evaluation Metrics}
\begin{figure*}
\begin{center}
   \includegraphics[width=\textwidth]{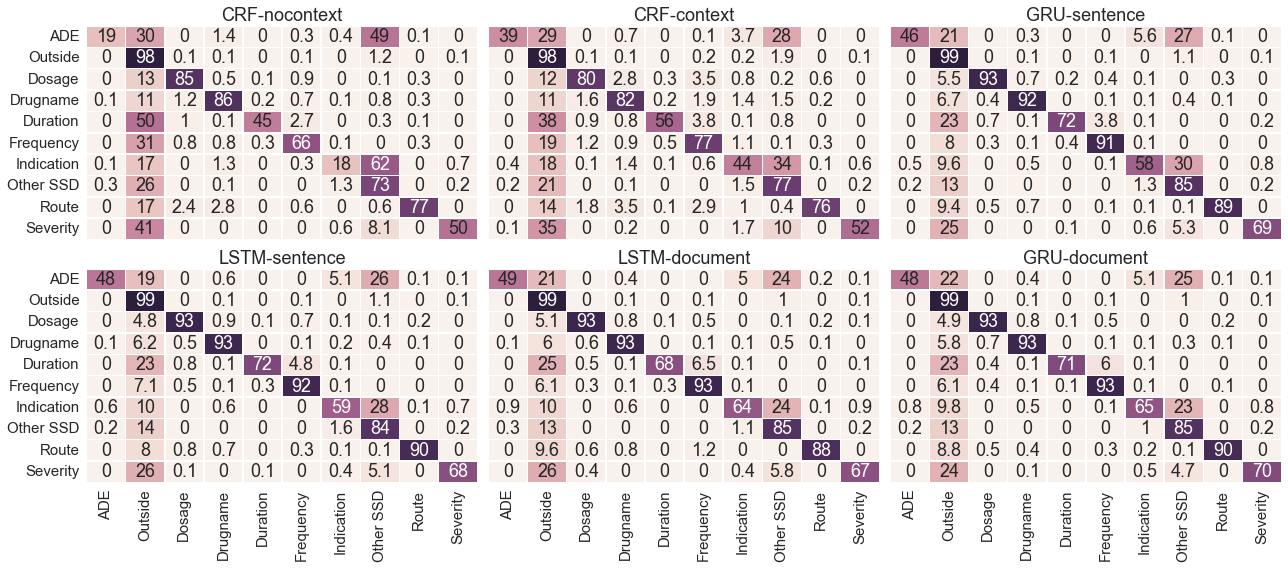}
\end{center}
   \caption{Heat-maps of Confusion Matrices of each method for the different class Labels. Rows are reference and columns are predictions. The value in cell $(i,j)$ denotes the percentage of words in label $i$ that were predicted as label $j$.}
\label{fig:loss}
\end{figure*}

For each word, the models were trained to predict either one of the nine medically relevant tags described in section 3, or the \textit{Outside} label. The CRF tagger was run in two modes. The first mode (CRF--– nocontext) used only the current word and its corresponding skip-gram representation. The second mode (CRF--– context) used the extra context feature described in section 4.3. The extra features are basically the bag of words representation of the preceding and following sections of the sentence. The first mode was used to compare the performance of CRF and RNN models when using the same input data. It also serves as a method of contrasting with CRF's performance when context features are explicitly added. CRF Tagger uses L-BFGS optimizer with L2- regularization.

The RNN frameworks are trained on sentence level and document level. The sentence level neural networks are fed only one sentence at a time. This means that the LSTM and GRU states are only preserved and propagated within a sentence. The networks cell states are re-initialized before each sentence.  The document level neural networks are fed one document at a time, so they can learn context cues that reside outside of the sentence boundary. We use 100 dimensional hidden layer for each directional RNN chain. Since we use bi-directional LSTMs and GRUs, this essentially amounts to a 200 dimensional recurrent hidden layer. The hidden layer activation functions for both RNN models are $tanh$. Output of this hidden layer is fed into a Softmax output layer which emits probabilities for each of the nine medical labels and the \textit{Outside} label. We use categorical cross entropy as the objective function. Similar to the CRF implementation, the Neural Net cost function also contains an L2-regularization component. We also use dropout  ~\cite{srivastava_dropout:_2014} as an additional measure to avoid over-fitting. Fifty percent dropout is used to manipulate the inputs to the RNN and the Softmax layer. We use AdaGrad ~\cite{duchi_adaptive_2011} to optimize the network cost.

We use ten-fold cross validation to calculate the performance metric for each model. The dataset is divided at the note level. We separate out 10 \% of the training set to form the validation set. This validation set is used to evaluate the different parameter combinations for CRF and RNN models. We employ early stopping to terminate the training run if the validation error increases consistently. We use a maximum of 40 epochs to train each network. The batch sizes used were kept constant at 128 for sentence level RNNs and 16 for document level RNNs. 

We report micro-averaged recall, precision and f-score. We use exact phrase matching to calculate the evaluation score for our experiments. Each phrase labeled by the learned models is considered a true positive only if it matches the exact true boundary of the phrase and correctly labels all the words in the phrase.

We use CRFsuite ~\cite{okazaki_crfsuite:_2007} for implementing the CRF tagger. We use Lasagne to setup the Neural Net framework. Lasagne\footnote{https://github.com/Lasagne/Lasagne} is a machine learning library focused towards neural networks that is build on top of Theano~\cite{bergstra_theano:_2010}.

\section{Results}
Table 2 shows the micro averaged scores for each method. All RNN models significantly outperform the baseline (CRF-context). Compared to the baseline system, our best system (GRU-document) improved the recall (0.8126), precision (0.7938) and F-score (0.8031) by 19\% , 2\% and 11 \%  respectively. Clearly the improvement in recall contributes more to the overall increase in system performance. The performance of different RNN models is almost similar, except for the GRU model which exhibits an F-score improvement of at least one percentage point over the rest.

The changes (gain or loss) in label wise F-score for each RNN model relative to the baseline CRF-context method are plotted in Figure 2. GRU-document exhibits the highest gain overall in six of the nine tags: \textit{indication} or diagnosis, \textit{route}, \textit{duration}, \textit{severity}, \textit{drug name}, and \textit{other SSD}. For indication, its gain is about 0.19, a near 50\% increase over the baseline. While the overall system performance of GRU-sentence, LSTM-sentence and LSTM-document are very similar, they do exhibit somewhat varied performance for different labels. The sentence level models clearly outperform the document level RNNs (both GRU and LSTM) for \textit{ADE} and \textit{Dosage}.  Additionally, GRU sentence model shows the highest gain in \textit{ADE} f-score.

Figure 3 shows the word level confusion matrix of different models for each label. Each cell shows the percentage of word tokens in row label $i$ that were classified as column label $j$. The consistent increase of diagonal entries of RNN models for all ten labels, indicates an increase in the overall system accuracy when compared to the baseline. The most densely populated column in this figure is the \textit{Outside} column, which denotes percentage of words that were erroneously labeled as \textit{Outside}.

Figure 4 shows the change in average F-scores for each method with changing percentage of training data used. The setup for training, development and test data is kept the same as the ten-fold cross validation setup mentioned in Section 5. Only the training data is randomly down-sampled to achieve the reduced training data size. The figure shows that Recurrent Neural Network models perform better than traditional CRF models even with smaller training data sizes.

\section{Discussion}
We already discussed in the previous section how improved recall seems to be the major reason behind improvements in the RNN F-score. This trend can be observed in Figure 3 where RNN models lead to significant decreases in confusion values present in \textit{Outside} column. 

Further examination of Figure 3 shows two major sources of error in the CRF systems. The largest source of error is caused by confusing the relevant medical words as \textit{Outside} (false negatives) and vice versa (false positives). The extent of false positives is not clear from Figure 3, but can be estimated if one takes into account that even a 1 \% confusion in the \textit{Outside} row represents about 5000 words. The second largest source of error is the confusion among \textit{ADE}, \textit{Indication} and \textit{Other SSD} labels. As we discuss in the following paragraphs, RNNs manage to significantly reduce both these type of errors.

The large improvement in recall of all labels for RNN models seems to suggest that RNNs are able to recognize a larger set of relevant patterns than CRF baselines. This supports our hypothesis that learning dependencies with variable context ranges is crucial for our task of medical information extraction from EHR notes. This is also evident from the reduced confusion among \textit{ADE}, \textit{Indication} and \textit{Other SSD}. Since these tags share a common vocabulary of Sign, Symptom and Disease Names, identifying the underlying word or phrase is not enough to distinguish between the three. Use of relevant patterns from surrounding context is often needed as a discriminative cue. Consequently,  \textit{ADE}, \textit{Indication} confusion values in the \textit{Other SSD} column for RNNs exhibit significant decreases when compared to CRF-nocontext and CRF-context.
\begin{figure}[t]
\begin{center}
   \includegraphics[width=0.47\textwidth]{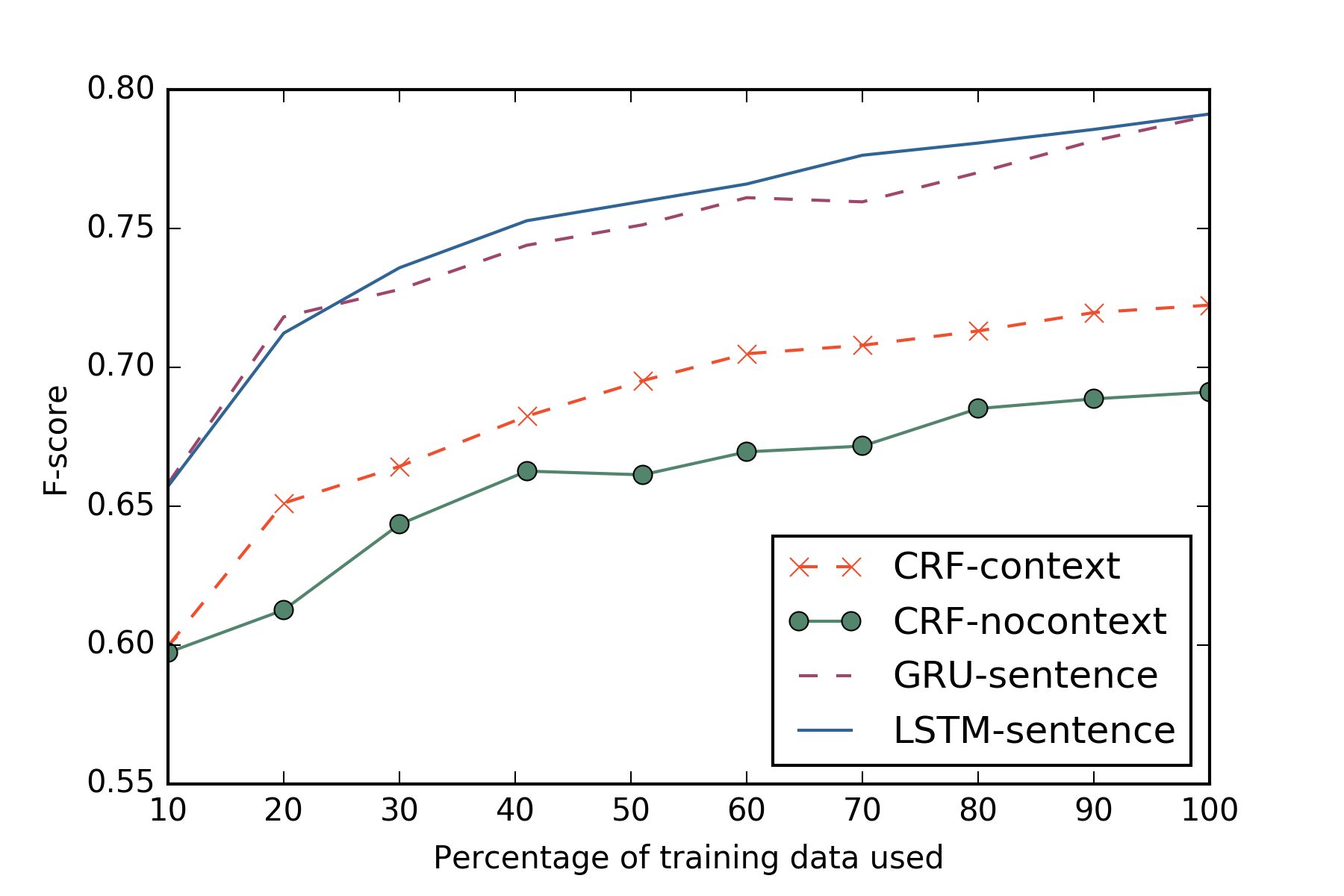}
\end{center}
   \caption{Change in F-score for all sentence models with respect to increasing training data size.}
\label{fig:datalength-model}
\end{figure}
We also see large improvements in detecting \textit{Duration}, \textit{Frequency} and \textit{Severity}. The vocabulary of these labels often lack specific medical jargon terms. Examples of these labels include ``seven days'', ``one week'' for \textit{duration}, ``some'', ``small'', ``no significant'' for \textit{severity} and ``as needed'', ``twice daily'' for \textit{frequency}. Therefore, they are most likely to be confused with \textit{Outside} label. This is indeed the case, as they have the highest confusion values in the \textit{Outside} column of CRF-nocontext. Including context in CRF improves the performance, but not as much as RNN models which decrease the confusion by almost half or more in all cases. For example, GRU-document only confuses Frequency as an unlabeled word about 6.1 \% of the time as opposed to 31 \% and 19 \% for CRF-nocontext and CRF-context respectively. 

Document level models benefit by using context from outside the sentence. Since the label \textit{Indication} requires the most use of surrounding context, it is clear that its performance would improve by using information from several sentences. Indications are diseases that are diagnosed by the medical staff, and the entire picture of the diagnosis is usually distributed across multiple sentences. Analysis of ADE is more complicated. Several ADE instances in a sentence also contain explicit cues similar to ``secondary to'' and ``caused by''. When coupled with \textit{Drugnames} this is enough to classify the ADE. Sentence level models might depend more on these local cues which leads to improved performance. Document models, on the other hand, have to recognize patterns from a larger context, using a very small dataset (total ADE annotations are just 905) which is quite difficult.

The LSTM-document model does not show the same improvement over the sentence models as GRU-document. One possible reason for this might be the simpler recurrence structure of GRU neuron as compared to LSTM. Since there are only 780 document sequences in the dataset, the GRU model with a smaller number of trainable parameters might learn faster than LSTM. It is possible that with a larger dataset, LSTM might perform comparable to or better than GRU. However, our experiments with reducing the hidden layer size of LSTM-document model to control for the number of trainable parameters did not produce any significant improvements. 

Moreover, figure 4 seems to indicate that there is not much difference between the performances of LSTM and GRU with different data sizes. However it is clearly surprising that RNN models with a larger number of parameters can still perform better than CRF models on smaller dataset sizes. This might be because the embedding layer, which contributes to a very large section of the trainable parameters, is initialized with a suitably good estimate using skip-gram word embeddings described in section 4.4. 

\section{Conclusion}

We have shown that RNNs models like LSTM and GRU are valuable tools for extracting medical events and attributes from noisy natural language text of EHR notes. We believe that the significant improvement provided by gated RNN models is due to their ability to remember information across different range of dependencies as and when required. As mentioned previously in the introduction, this is very important for our task because different labels have different contextual dependencies. CRF models with hand crafted features like bag of words representation, use fixed context windows and lose a lot of information in the process. 

RNNs are excellent in extracting relevant patterns from sequence data. However, they do not explicitly enforce constraints or dependencies over the output labels. We believe that adding a probabilistic graphical model framework for structured output prediction would further improve the performance of our system. This experiment remains as our future work.

\section*{Acknowledgments}
We thank the UMassMed annotation team, including Elaine Freund, Wiesong Liu and Steve Belknap, for creating the gold standard evaluation set used in this work.  We also thank  the anonymous reviewers for their comments and suggestions.

This work was supported in part by the grant 5U01CA180975 from the National Institutes of Health (NIH). We also acknowledge the support from the United States Department of Veterans Affairs (VA) through Award 1I01HX001457. The contents of this paper do not represent the views of the NIH, VA or United States Government.  
\bibliography{refv1}
\bibliographystyle{naaclhlt2016}

\end{document}